%% file: root.tex
\documentclass[letterpaper, 10 pt, journal, twoside]{ieeetran}
\makeatletter
\let\NAT@parse\undefined
\makeatother
\input{math_commands.tex}

\usepackage[numbers,]{natbib}
\usepackage{amsmath}
\usepackage{amssymb}
\usepackage{booktabs}
\usepackage{multirow}
\usepackage{graphicx}
\usepackage{paralist}
\usepackage{wrapfig}
\usepackage{lipsum}
\usepackage{stfloats}
\usepackage{algorithm}
\usepackage{algpseudocode}
\usepackage{hyperref}

\usepackage[table]{xcolor}
\usepackage{xcolor}
\usepackage{url}
\usepackage{caption}
\usepackage{subcaption}
\usepackage{amsmath}
\usepackage{graphicx} 
\newtheorem{lemma}{Lemma}
\newtheorem{proposition}{Proposition}
\usepackage{float}
\usepackage{enumerate}
\usepackage{array}
\usepackage{subfiles}
\usepackage{anyfontsize}

\newcommand{\SE}{\mathrm{SE}}
\newcommand{\SO}{\mathrm{SO}}
\newcommand{\pick}{\mathrm{pick}}
\newcommand{\place}{\mathrm{place}}
\newcommand{\out}{\mathrm{out}}

\usepackage{pifont}

\newcommand{\p}{\mathrm{pick}}
\newcommand{\q}{\mathrm{place}}

\newcommand{\eref}{Equation~\ref}

\hypersetup{hidelinks}

\begin{document}
\bstctlcite{BSTcontrol}

\title{Learning Efficient and Robust Language-conditioned Manipulation using Textual-Visual Relevancy and Equivariant Language Mapping}

\author{Mingxi Jia$^{\dagger1}$, Haojie Huang$^{\dagger2}$, Zhewen Zhang$^{\ddagger2}$, Chenghao Wang$^{\ddagger2}$, Linfeng Zhao${^2}$, \\Dian Wang${^2}$, Jason Xinyu Liu${^1}$, Robin Walters${^2}$, Robert Platt$^{*2}$, and Stefanie Tellex$^{*1}$
\thanks{Manuscript received: January, 29, 2025; Revised April, 30, 2025; Accepted June, 5, 2025.}
\thanks{This paper was recommended for publication by Editor Markus Vincze upon evaluation of the Associate Editor and Reviewers' comments.
This work is supported by ONR under grant numbers N000142412784.} 
\thanks{$\dagger, \ddagger$ Equal Contribution, $\star$ Equally Advising}
\thanks{$^{1}$Mingxi Jia, Jason Xinyu Liu, and Stefanie Tellex are with Brown University, Providence, RI, USA
        {\tt\footnotesize \{mingxi\_jia, xinyu\_liu, stefanie\_tellex\}@brown.edu}}%
\thanks{$^{2} $Haojie Huang, Zhewen Zhang, Chenghao Wang, Linfeng Zhao, Dian Wang, Robin Walters, and Robert Platt are with Northeastern University, Boston, USA
        {\tt\footnotesize \{huang.haoj, zhang.zhew, wang.chengh, zhao.linf, wang.dian, r.walters\}@northeastern.edu, rplatt@ccs.neu.edu}}%
\thanks{Digital Object Identifier (DOI): see top of this page.}
}

\markboth{Journal of \LaTeX\ Class Files,~Vol.~14, No.~8, August~2015}%
{Shell \MakeLowercase{\textit{et al.}}: Bare Demo of IEEEtran.cls for IEEE Journals}
\markboth{IEEE Robotics and Automation Letters. Preprint Version. June, 2025}
{Jia \MakeLowercase{\textit{et al.}}: GEM}

\maketitle

\begin{abstract}

Controlling robots through natural language is pivotal for enhancing human-robot collaboration and synthesizing complex robot behaviors. Recent works that are trained on large robot datasets show impressive generalization abilities. However, such pretrained methods are (1) often fragile to unseen scenarios, and (2) expensive to adapt to new tasks.  This paper introduces \textbf{G}rounded \textbf{E}quivariant \textbf{M}anipulation (\textbf{GEM}), a robust yet efficient approach that leverages pre-trained vision-language models with equivariant language mapping for language-conditioned manipulation tasks. Our experiments demonstrate GEM's high sample efficiency and generalization ability across diverse tasks in both simulation and the real world. GEM achieves similar or higher performance with orders of magnitude fewer robot data compared with major data-efficient baselines such as CLIPort and VIMA. Finally, our approach demonstrates greater robustness compared to large VLA model, e.g, OpenVLA, at correctly interpreting natural language commands on unseen objects and poses. Code, data, and training details are available \url{https://saulbatman.github.io/gem_page/}

\end{abstract}

\begin{IEEEkeywords}
Deep Learning in Grasping and Manipulation, Learning from Demonstration
\end{IEEEkeywords}

\section{INTRODUCTION}

\IEEEPARstart{C}{ommanding} a robotic manipulator with natural language is important for enabling human-robot collaboration, while adding language specifications increases the complexity of the policy learning problem because the model needs to deal with a higher dimensional state space, especially in 
the multi-task setting. Data-driven approaches open new possibilities for learning fine-grained language-conditioned policies by learning from human demonstrations. Yet, these methods require extensive data collection~\cite{shridhar2022cliport, jiang2022vima, brohan2022rt} or expensive finetuning~\cite{kim2024openvla, brohan2023rt}. For example, CLIPort~\cite{shridhar2022cliport} needs 50-100 demonstrations per task to ensure robust performance. RT-2~\cite{brohan2023rt} finetunes a 55B LLM model with 130k robot trajectories and 10 billion images-text pairs. 

Another strand of prior works~\cite{liu2024moka, huang2023voxposer, mirchandani2023large, liang2023code} loosens the need for large robotic datasets by directly utilizing existing Vision Language Models (VLM) to generate robot commands. However, since the VLMs lack an understanding of the geometries of the physical world, these approaches are still fragile to unseen object poses and cannot perform accurate, fine-grained manipulation on novel shapes.

\begin{figure}[t]
    \centering
    \includegraphics[width=0.8\linewidth]{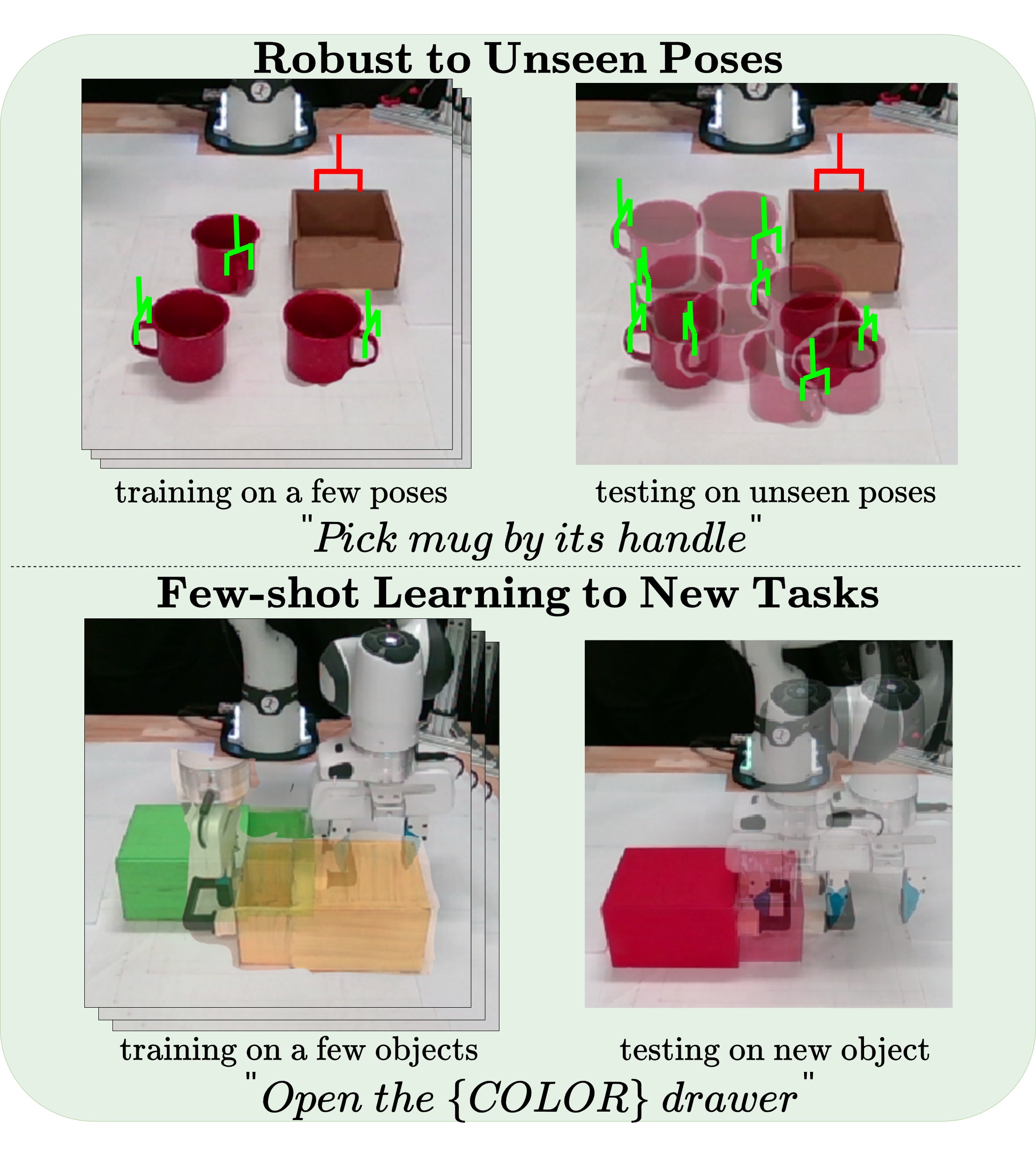}
    \vspace{-10pt}
    \caption{
  \small
  \textbf{Overview.} By leveraging textual-visual relevancy maps with equivariant language mapping, it allows the robot to learn robust manipulation policies from a few demonstrations with high data efficiency. Our method achieves, in average, 10x data efficiency against CLIPort~\cite{shridhar2022cliport} and 53\% more robustness than OpenVLA~\cite{kim2024openvla}.
  }
  \label{fig:intro_example}
  \vspace{-15pt}
\end{figure}

Therefore, two key weaknesses of prior work are (1) data-driven methods require extensive data collection and expensive fine-tuning if a new task is to be learned due to the complex state space of potential objects and poses as well as the large action space of possible robot movements, and (2) the zero-shot approches are fragile to unseen object poses and distractor objects. We address these challenges by introducing a new imitation learning algorithm that leverages textual-visual relevancies from pretrained VLMs and equivariant language mapping that enables our method to learn efficient yet robust language-conditioned policies.  

Our approach, \textbf{G}rounded \textbf{E}quivariant \textbf{M}anipulation (\textbf{GEM}) exploits domain symmetries that exist in the language-conditioned manipulation problem, specifically, equivariance in $\SE(2)$. For example, consider ``grasp the coffee mug by its handle.'' If there is a rotation or translation of the mug, the desired pick action should also transform accordingly, i.e., equivariantly. Our learned policy incorporates such language-conditioned symmetries and generalizes to unseen scenarios related by symmetric transformations, thus making the model generalize to unseen object poses in a few-shot manner. To enable novel object generalization across objects, colors, and shapes, we propose textual-visual relevancy maps to distill information from large vision and language models. We extract textual relevancy for open-vocabulary recognition and construct visual relevancy via data retrieval to improve robustness. We leverage spatial action map~\cite{zeng2021transporter}, i.e., representing actions as pixel locations, with open-loop action primitives that allow us to flexibly leverage equivariance and semantic relevancies. The resulting method is robust to unseen object poses and achieves better novel object generalization ability, as shown in Figure~\ref{fig:intro_example}.

We make the following specific contributions. \textbf{(1)} We propose a novel method to construct patch-level textual and visual relevancy maps for robust open-vocabulary generalization. \textbf{(2)} We analyze the symmetries underlying the language-conditioned manipulation problem and design a novel neural network called language steerable kernels for building language-conditioned equivariant networks. \textbf{(3)} We demonstrate the state-of-the-art generalization robustness and sample efficiency in both simulation and the real world on a series of challenging language-conditioned manipulation tasks, using only $10\%$-$20\%$ training data compared with CLIPort~\cite{shridhar2022cliport} and $0.1\%$ training data compared with VIMA~\cite{jiang2022vima}. 

\section{Related work}
\label{sec:related}

\textbf{Language-conditioned Policy Learning:}
Using language as task specifications is a common way for multi-task policy learning.  Prior works~\cite{shao2021concept2robot, tziafas2023language, tang2023task, jiang2022vima,goyal2023rvt} use feature concatenation, FiLM~\cite{perez2018film}, or the self/cross-attention mechanism to fuse image and language features. For example, \citet{shridhar2022cliport} utilizes CLIP visual and text encoders and fuses language and image features with a two-stream architecture. \citet{goyal2023rvt} processes the language token and visual token jointly with a transformer. These models treat language as static features, meaning that the language features remain the same no matter how visual features change in a specific task. These designs require lots of data to cover all possible variances of visual features to guarantee performance. For instance, VIMA~\cite{jiang2022vima} requires 60k robot demos to learn its visual pick \& place tasks. In contrast, our method maps language instructions into equivariant features that dynamically apply over the entire image space to generate robot actions, allowing our method to achieve similar or higher success rates with only 10-20 demonstrations.

\textbf{Data-efficient Few-shot Learning} requires the robot to learn a robust policy to manipulate in-distribution objects given only a few demonstrations. Prior works using geometric-constrained models~\cite{zeng2021transporter, huang2022equivariant, wang2022so, huang2023edge, hu2024orbitgrasp, zhao2023e2equivariant, simeonov2022neural, yang2023equivact, zhao2023integrating, wangequivariant} leverage symmetries in robotic tasks and have demonstrated their superior effectiveness for unseen pose generalization that allows few-shot learning. However, these methods often learn a single-task policy and only take visual inputs. It is challenging to incorporate language into equivariant models because, unlike images, language does not have explicit spatial properties like rotation or translation. In this paper, we propose a novel architecture that treats language as a dynamic feature and enables object-level equivariance using equivariant language mapping via Language Steerable Kernels. Our approach can learn a multi-task policy yet maintain high sample efficiency and spatial robustness.

\textbf{Zero-shot Manipulation} requires the robot to generalize to out-of-distribution objects with novel shapes, colors, or textures. Learning-from-scratch methods~\citep{zeng2021transporter,florence2019self,wang2023mimicplay} perform well on seen objects but cannot generalize well on unseen ones. Prior work~\cite{lerftogo2023,ahn2022can,hu2023look} uses LLM/VLMs as a zero-shot object detector or a text-level task planner, assuming access to skills like a grasping model and pose estimators. However, the lack of learning ability limits these methods from adapting to complex task-oriented behaviors, e.g., ``insert the letter E block into the letter E hole'' since there is no general actor available for such placing skills. F3RM~\cite{shen2023distilled} learns few-shot pick-and-place with distilled fields, but it is slow because it requires extensive camera views and test-time optimization for every action step. RT-2~\citep{brohan2023rt} and OpenVLA~\citep{kim2024openvla} learn generalist vision-language-action models by training on large robotic datasets, which give a certain degree of emerging behaviors on unseen objects with novel instructions. Still, it is expensive to collect new datasets and finetune these large parameterized models to adapt to new tasks. In this paper, we propose a novel approach that is capable of learning policies in a predefined space of skills with a small number of demonstrations while leveraging the zero-shot generalization ability from pre-trained VLM models via patch-level textual and visual relevancy maps.

\section{Method}

\label{sec:method}
\textbf{Problem Statement and Assumptions:} We consider the problem of learning from demonstrations for language-conditioned manipulation tasks in $\SE(2)$ space. We frame it as a two-step pick-and-place learning problem, although our approach can support a larger space of skills, including pushing, pulling, etc. Given a set of demonstrations that contains observation-language-action tuples $(O_t,\ell_t,a_t)$, the objective is to learn a policy $p(a_t|o_t,\ell_t)$ where the action $a_t = (a^{\p}_t,a^{\q}_t)$ has pick and place components. Please note the policy can be formulated to
generate the multi-step pick-place actions, and we discard time step subscripts $t$ for simplicity in the following sections. 
The visual observation $O_t$ is a set of RGBD images at several camera views. The two-step actions, $a^{\p}$ and $a^{\q}$, are parameterized in terms of $\SE(2)$ coordinates $(u,v,\theta_{\p})$ and $(u,v,\theta_{\q})$, respectively, where $u,v$ denotes the pixel coordinates of the gripper position, $\theta_{\p}$ is the pick orientation defined with respect to the world frame, and $\theta_{\q}$ is the delta angle between the pick and place. 
The language instruction $\ell_t$ specifies the current-step instruction, e.g., ``open the drawer'' or ``grasp scissors by its handle and place into box.'' We assume $\ell_t$ for each step can be parsed into the pick instruction and the place instruction, $\ell = (\ell^{\p},\ell^{\q})$.

\textbf{Method Overview:} There are three main modules. \textbf{(1)} The relevancy extraction module takes multi-view images $\mathcal{O}$ and the language instruction $\ell$. It outputs a dense relevancy map that summarizes the visual and language input. \textbf{(2)} The language-conditioned pick module takes as input the top-down orthographic RGB-D projection $o$ of the scene with the language instruction $\ell$ and outputs an action map over pick actions. \textbf{(3)} Similarly, the language-conditioned place module produces an action map over place actions. The difference is that the place module does convolution with an image-crop-conditioned kernel instead of a language-conditioned kernel. 

\begin{figure*}[t]
    \vspace{5pt}
    \centering
    \includegraphics[width=0.87\textwidth]{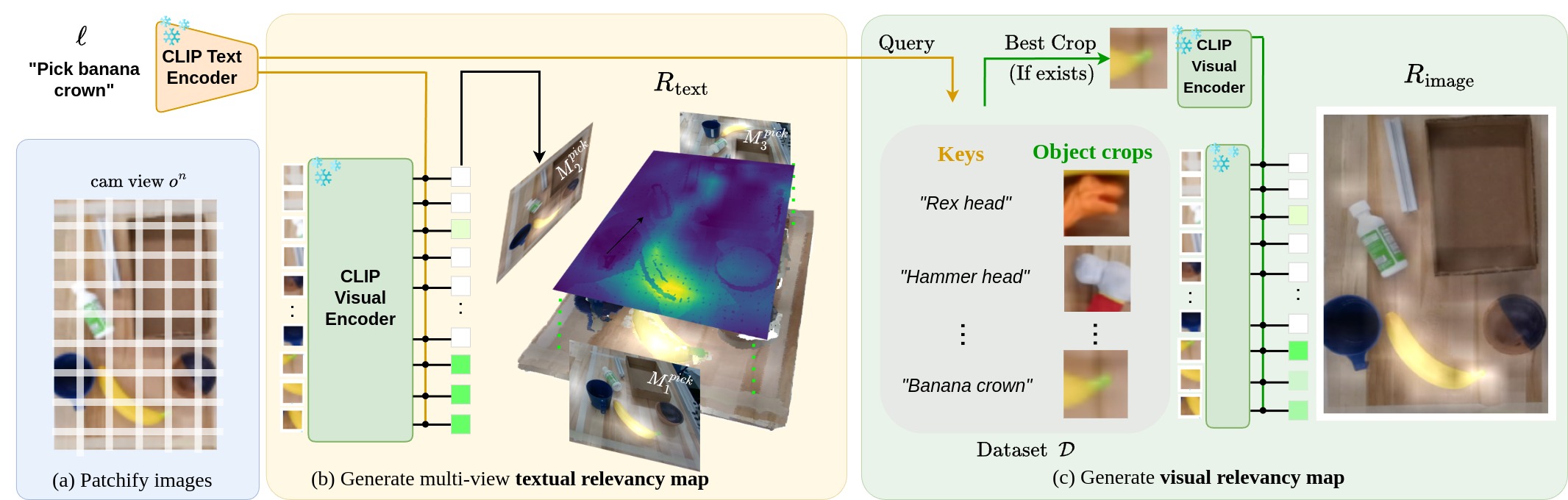}
    \caption{
    \small
    \textbf{Relevancy Map Extraction.} After (a) \textbf{patchifying images} into patches, we compute two types of relevancy maps. (b) \textbf{Textual Relevancy Maps} allow open-vocabulary recognition, which is a topdown projection from the relevancy point cloud constructed with multi-view relevancy maps. We reconstruct the point cloud by reprojecting RGBD with camera intrinsics and merging different views with extrinsics. (c) \textbf{Visual Relevancy Maps} enhance the recognition robustness by querying from the existing database. }
    
    \vspace{-18pt}
    \label{fig:semantic}
    
\end{figure*}
\vspace{-0.1cm}
\subsection{Zero-shot Recognition via Textual \& Visual Relevancies}
\label{sec:img-cond-semantic}
The relevancy extraction module uses a pre-trained CLIP model to identify parts of the visual observation most relevant to the current task. Specifically, it takes the language goals $\ell^{\p}$ and $\ell^{\q}$ and the current $N$-view observations $\mathcal{O} = \{o^1, o^2, ..., o^N\}$ as input and produces relevancy maps $R^{\p}$ and  $R^{\q}$ that highlight the language goals in the pixel space. Note that while these relevancy maps do not tell the system exactly where to pick, they provide a strong visual-language prior. The pipeline is illustrated in Figure~\ref{fig:semantic}.

     \textbf{Textual Relevancy Maps for Zero-shot Recognition:} We use CLIP~\citep{radford2021learning}, which was trained by minimizing the cosine similarity between the image feature and its text label with internet data, to generate a pixel-wise relevancy score for each of the $N$ views in $\mathcal{O}$. We split each image along a grid into image patches with patch size $p$ and stride $s$. Each RGB image patch is then scored with its cosine similarity to the language instructions with pre-trained CLIP features. The textual relevancy function $\mathcal{R}_{text}$ can be described by
     \begin{equation}
        \label{eqn:text-map}
        \mathcal{R}_{\mathrm{text}}(\mathcal{P}(o^n), \ell) = \mathcal{P}^{-1}(\mathcal{E}_{\mathrm{patch}} \cdot \mathcal{E}_{\ell}^T),
      \end{equation} 
      where $\mathcal{P}$ denotes the image patchification function and $\mathcal{P}^{-1}$ denotes an inverse process that transforms all similarity scores back to the original image dimension. $\mathcal{E}_{patch}\in \mathbb{R}^{(m\times n)\times d_m}$ is the embedding outputs from the CLIP image encoder, where $(m\times n)$ and $d_m$ denote the number of image patches and the output embedding dimension of CLIP respectively. $\mathcal{E}_{\ell}\in \mathbb{R}^{1\times d_m}$ is the embedding output for language instruction $\ell$ from the CLIP text encoder. After getting pixel-wise relevancies for each of the $N$ views, we integrate this information into a single point cloud and label each point with the corresponding relevancy maps from all views so that we get a final top-down textual relevancy map $R_{text}$ via projection, as shown in panel (b) of Figure~\ref{fig:semantic}.

     \begin{figure}[t]
    \centering
    \includegraphics[width=0.8\linewidth]{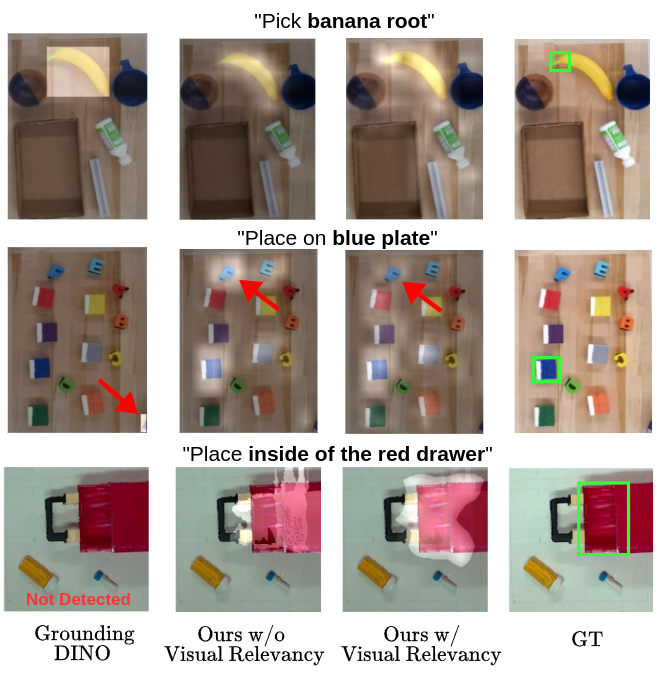}
    \vspace{-5pt}
    \caption{
    \small
    \textbf{Qualitative Results of Relevancy Maps.} Our recognition accuracy with visual relevancy focuses more on the task-specific part (first-row), is less distracted by color (second-row), and recognizes uncommon instructions (third-row) than the baseline~\cite{liu2023grounding}. The text-visual ratio is set to 0.8 for visualization.}
    \label{fig:visualization}
    \vspace{-25pt}
    \end{figure}

    \textbf{Visual Relevancy Map for Semantic Correction:} Although the textual relevancy map highlights image regions related to language instructions, we found it insufficient because the high-value region does not necessarily highlight the correct object if certain text is underrepresented during CLIP training. This misalignment creates noisy training samples as shown in Figure~\ref{fig:visualization} where \textit{Grounding DINO} only highlights a rough area and \textit{Ours w/o Visual Relevancy} is biased by colors. To solve it, we further introduce the visual relevancy map, which is illustrated in Figure~\ref{fig:semantic}(c). Starting with the demonstrations, we identify the image crops in the demonstration data corresponding to pick and place events. For all pick/place events identified, we store into a database a pair comprised of the image patch at the pick/place location and the language query that describes the pick/place object (left side of Figure~\ref{fig:semantic}(c)). Then, at inference time, we index into the dataset using the language query text and recall the corresponding image crop, e.g., recall the image crop from the dataset corresponding to ``banana root''. The crop query process can be expressed by
      \begin{equation}
        \label{eqn:text-query}
        \text{QueriedCrop}(\mathcal{D}, \ell) = \argmax_{\mathrm{crop}\in \mathcal{D}} (\mathcal{E}_\ell \cdot K_{\mathrm{crop}}^T),
      \end{equation} 
     where $K_{\mathrm{crop}}\in \mathbb{R}^{N \times d_m}$ denotes all language embeddings that correspond with $N$ image patches for all $N$ pick and place objects in dataset $\mathcal{D}$. $\mathcal{E}_\ell \in \mathbb{R}^{1 \times d_m}$ denotes the embedding of the language query. Then, we generate visual relevancy maps by evaluating the cosine similarity between the CLIP embeddings of the recalled image crop and the patch embeddings from the top-down image $o_t$. The visual relevancy generation function $\mathcal{R}_{\mathrm{visual}}$ can be described by
     \begin{equation}
        \label{eqn:image-map}
        \mathcal{R}_{\mathrm{visual}}(\mathcal{P}(o), \ell, \mathcal{D}) = \mathcal{P}^{-1}(\mathcal{E}_{\mathrm{patch}} \cdot \mathcal{E}_{\mathrm{crop}}^T),
      \end{equation} 
    where $\mathcal{E}_{\text{crop}}\in \mathbb{R}^{1\times d_m}$ denotes the embedding output for the image crop from the CLIP image encoder. If the pick/place target cannot be located in the dataset, i.e., $\max(\mathcal{E}_\ell \cdot K_{\mathrm{crop}}^T)$ is below a threshold, then the visual relevancy map function returns None. See Figure~\ref{fig:visualization} for visualization examples. 

     To fuse textual and visual relevancy maps, we do a pixel-wise weighted averaging if a \textit{Best Crop} exists in the dataset. Otherwise, only the textual map will be used.

\begin{figure}[t]
    \centering
    \vspace{10pt}
    \includegraphics[width=0.75\linewidth]{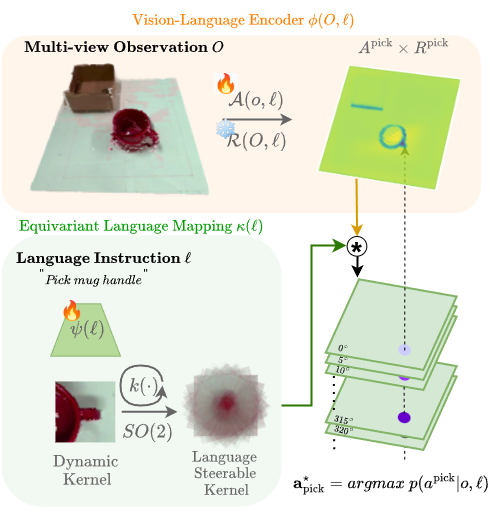}
    \caption{
    \small
    \textbf{Object-level $\SE(2)$-equivariant Picking Model.} Our picking model $f^\p$ consists of two branches. The top branch is our vision-language encoder $\phi(O, \ell)$, which includes a learnable action model $\mathcal{A}$ and the textual-visual relevancy extractor $\mathcal{R}$. The bottom branch is the Equivariant Language Mapping $\kappa$, which generates the language steerable kernel by rotating the dynamic kernel $\psi(\ell)$. The fire and the snowflake symbol denote frozen and trainable models. We discretize the $\SO(2)$ rotation space into $72$ bins.}
    \label{fig:pick}
    \vspace{-20pt}
\end{figure}

\subsection{Language Equivariance for Few-shot Learning:}
While the relevancy maps provide semantic guidance for recognizing objects, the robot has no knowledge of where to manipulate the object shapes. Here, we use Learn from Demonstrations (LfD) to encode human knowledge into a learning-based model. A desired property we want a language-conditioned policy model to have is object-level equivariance, meaning that the output action should transform accordingly as the object specified by the language instruction transforms in the workspace. In the following, we propose a novel neural network architecture that enables equivariant language Mapping with its mathematical proofs. 

\textbf{Equivariant Language Mapping:}  We propose Equivariant Language Mapping via \textit{language steerable kernels} that leverage object-level $\SE(2)$ equivariance, which allows our model to learn a new task in a few-shot learning manner. Previous work~\citep{zeng2021transporter, shridhar2022cliport} infers the pick location by taking the argmax on the output attention map from UNet, which has no guarantee that the model will generalize if any object moves in $\SE(2)$. Instead, we leverage the object-level symmetry by first mapping text embeddings by a UNet $\psi^{\p}(\ell^{\p})$ into a kernel with a spatial dimension of (C, H, W) and then converting it into steerable kernels by rotating the output with a group of $n$ rotations $\{\frac{2\pi i}{n}|  0\leq i < n\}$. This results in a stack of $n$ rotated feature maps, $\Psi(\cdot)=\{g_0\cdot\psi(\cdot), g_2\cdot\psi(\cdot),\cdots, g_{n-1}\cdot\psi(\cdot) \}$, where $g_i = \frac{2\pi i}{n}$. We define the entire process as a language steerable kernel generator $\kappa^{\p}(\ell^{\p})$. The language kernel $\Phi$ maps language embeddings to convolutional kernels that satisfy the steerability constraints~\citep{cesa2021program}. It allows picking action inference to be $\SO(2)$ equivariant with respect to the object poses.

\textbf{Symmetry of Language-Conditioned Pick:} The picking model $f^{\pick}$ calculates a probability distribution over gripper pose that corresponds to the probability of a successful grasp on the desired object part. This distribution $p(a^{\p} | o, \ell^{\p})$ is estimated by $f^{\p} (o,\ell^{\p},  M^{\p})$. The pick command executed by the robot is selected by
$a_{\pick}^{\star} = \argmax a^{\pick}$. The desired pick action is equivariant with respect to the pose of the object to be picked, i.e., $g \cdot p(a^{\p} | b^{\p}, \ell^{\p}) = p(a^{\p} | g \cdot b^{\p}, \ell^{\p})$, where $b^{\p}$ denotes the object to be picked and $g \cdot$ denotes the action of a transformation $g$. Note that this equivariance is local to the object, in contrast to standard models that are equivariant w.r.t. the scene. 

Specifically, assume the observation $o_t$ contains a set of $m$ objects $\{b_i\}_{i=1}^{m}$ on the workspace and denote the object $b^{\p}$ as the goal object instructed by the language instruction $\ell^{\p}$. If there is a transformation $g\in \SE(2)$ on the target object $b^{\p}$ regardless of transformations on other objects, we denote it as $g\cdot o^{b^{\p}}$. The symmetry underlying $f$ can be stated as 
\begin{equation}
\begin{split}
    \argmax f^{\p}(g \cdot o^{b^{\p}}, \ell^{\p}) &\\= 
     g\cdot \argmax f^{\p}(o, \ell^{\p})
    \label{eqn:pick-symmetries}
\end{split}
\end{equation}

\eref{eqn:pick-symmetries} claims that if there is transformation $g\in \SE(2)$ on the object $b^{\ell}$, the best action  ${a}^{\star}_{\p}$ to grasp the instructed object should be transformed to $g\cdot {a}^{\star}_{\p}$. If the symmetry is encoded in our pick model, it can generalize the knowledge learned from the demonstration to many unseen configurations. We use this symmetry to improve sample efficiency and the generalization of our pick model.

\textbf{Pick Model Architecture:} There are two main parts of the pick model. The first (shown in the top part of Figure~\ref{fig:pick}) calculates a language-conditioned pick map as follows. We feed the raw RGB-D observation into a UNet, denoted as $\mathcal{A}^{\p}$, and encode $\ell^{\pick}_t$ with the CLIP.
The encoded language vector is concatenated onto the descriptor of every pixel in the bottleneck layer of $\mathcal{A}^{\p}$. The output of $\mathcal{A}^{\p}$ is denoted as $A^{\p}(o,\ell^{\p})$ or $A^{\p}$ for simplicity. It is then integrated with the pick relevancy map $R^{\p}$ with element-wise multiplication, shown as $\otimes$ in Figure~\ref{fig:pick}.

\textbf{Symmetry of Language-Conditioned Place:}
Place action that transforms pick target to the placement are bi-equivariant~\citep{huang2024fourier,ryu2022equivariant}, i.e., independent transformations of the placement with $g_1$ and the pick target with $g_2$ result in a change ($a'_{\place}=g_1 a_{\place} g_2^{-1}$) to complete the rearrangement at the new configuration. Leveraging the bi-equivariant symmetries can generalize the place knowledge to different configurations and thus improve the sample efficiency.
The coupled symmetries exist in the language-conditioned place:
\begin{equation}
\begin{split}
    \argmax f^{\q}(g_1 \cdot o^{b^{\place}}+g_2 \cdot o^{b^{\p}}, \ell^{\place}) =&\\
    g_1 \theta(g_2^{-1}) \cdot \argmax f^{\q}(o, \ell^{\place})
    \label{eqn:place-symmetries-3}
\end{split}
\end{equation}
where $g_1 \cdot o^{b^{\place}}+g_2 \cdot o^{b^{\p}}$ denotes $g_1 \in \SE(2)$ and $g_2 \in \SE(2)$ acting on the instructed placement $b^{\q}$ and the picked object $b^{\p}$, respecitively. $\theta(g^{-1}_2)$ denote the angle of the place action is rotated by $-g_2$. Specifically, the RHS of Equation~\ref{eqn:place-symmetries-3} indicates that the best place location is rotated by $g_1$, and the place orientation is rotated by $\theta(g_1)\theta(g_2^{-1})$.
Our place model is designed to satisfy the language-conditioned equivariance of Equation~\ref{eqn:place-symmetries-3}.

\subsection{Equivariance Proof for Language Steerable Kernel}
Here, we prove the object-level $\SE(2)$ equivariance via language steerable kernels as stated in Equation~\ref{eqn:pick-symmetries}.

\textbf{Steeribility background:} The G-steerable kernels are convolution kernels $K\colon \mathbb{R}^{n} \rightarrow \mathbb{R}^{d_{\mathrm{out}}\times d_{\mathrm{in}}}$ satisfying the \emph{steerability constraint}, where $n$ is the dimensionality of the space, $d_{\mathrm{out}}$ and $d_\mathrm{in}$ are the output and input field type
\begin{equation}
    K(g\cdot x) = \rho_{\mathrm{out}}(g)K(x)\rho_{\mathrm{in}}(g)^{-1}
    \label{equ:steerablility_constraint}
\end{equation} 
\begin{proposition}
    If $\kappa(\ell)$ is a steerable kernel, it approximately satisfies the symmetry stated in Equation~\ref{eqn:pick-symmetries}. 
    \label{prop1}
\end{proposition}

\textbf{Translational Equivariance.} Since FCNs are translationally equivariant by their nature, if a target object $o^b$ is translated to a new location, the cross-correlation between $\kappa(\ell) \ast f^\p(o,\ell)$ will capture this translation, and there is no change in the change space.

\textbf{Rotation Equivariance.} Assuming $\phi$ satisfies the equivariant property that $\phi(T_g^0 o,\ell) = T_g^0 \phi(o,\ell)$ and the rotation of $o^b$ is represented by $T_g^{0} o_t$, we start the proof with lemma~\ref{lemma1} and lemma~\ref{lemma2}.

\begin{lemma}
    \label{lemma1}
    if $k(x)$ is a steerable kernel that takes trivial-type input signal, it satisfies ${T_g^{0}K(x) =  \rho_{\mathrm{out}}(g^{-1})K(x)}$.
\end{lemma}

\begin{lemma}
\label{lemma2}

Cross-correlation satisfies that
  \begin{align} 
    (T_g^0(K\star f))(\vec{v})= ((T_g^0 K) \star (T_g^0 f))(\vec{v})  
  \end{align}
\end{lemma}
  \noindent

Given Lemma~\ref{lemma1} and lemma~\ref{lemma2}, we can prove that 
\begin{align*}
    \kappa(\ell) \ast \phi(T_g^{0} o , \ell) =& \kappa(\ell) \ast T_g^{0}\phi(o , \ell)\\
    =& \kappa(\ell) \ast T_g^{0}\phi(o , \ell)\\
    =& T_g^{0} T_{g^{-1}}^{0}\kappa(\ell) \ast T_g^{0}\phi(o , \ell)\\
    =& T_g^{0} [T_{g^{-1}}^{0}\kappa(\ell) \ast \phi(o , \ell)] \:\:\text{lemma 2}\\
    = & T_g^{0} [\rho_{\out}(g)\kappa(\ell) \ast \phi(o , \ell)]\:\: \text{lemma 1}
\end{align*}

It states that if there is a rotation on $o$, the grasp position is changed by $T_g^{0}$, and the rotation is changed by $\rho_{\out}(g)$. Since the cross-correlation is calculated for each pixel without stride, the rotated $b^{\ell}$ is captured by $\rho(g)$. In our implementation, we generate the language-conditioned steerable kernel $\kappa(\ell)$ but remove the constraint of the equivariant property of $\phi$. The U-Net encoder maintains soft global equivariance with the long skip connections and data augmentation.


\begin{table*}[t!]
\vspace{5pt}
  \setlength\tabcolsep{2pt}
  \centering
  \scriptsize
    \begin{tabular}{@{}l*{22}{>{\centering\arraybackslash}p{8mm}@{}}}
    \toprule
    & \multicolumn{3}{c}{\begin{tabular}[c]{@{}c@{}}packing-box-pairs\\seen-colors\end{tabular}}    & \multicolumn{3}{c}{\begin{tabular}[c]{@{}c@{}}packing-box-pairs\\unseen-colors\end{tabular}}    & \multicolumn{3}{c}{\begin{tabular}[c]{@{}c@{}}packing-seen-google\\objects-seq\end{tabular}} & 
    \multicolumn{3}{c}{\begin{tabular}[c]{@{}c@{}}packing-unseen-google\\objects-seq\end{tabular}} & 
    \multicolumn{3}{c}{\begin{tabular}[c]{@{}c@{}}packing-seen-google\\objects-group\end{tabular}} & 
    \multicolumn{3}{c}{\begin{tabular}[c]{@{}c@{}}packing-unseen-google\\objects-group\end{tabular}}  \\ [-2pt]
    
    \cmidrule(lr){2-4} \cmidrule(lr){5-7} \cmidrule(lr){8-10} \cmidrule(lr){11-13} \cmidrule(lr){14-16} \cmidrule(lr){17-19} \cmidrule(lr){20-22}
        Model &  \multicolumn{1}{c}{10}  & 20   & 100 & \multicolumn{1}{c}{10}  & 20   & 100 & 10 & 20 & 100 &  10 & 20 & 100 & 10  &  20 & 100 & 10 & 20 & 100 & \\
 \midrule
     
    Transporter-Lan~\citep{zeng2021transporter} & 50.4 & 72.4 & 86.8 & 41.2 & 40.4 & 60.7 & 36.3 & 57.0 & 83.2 & 31.7 & 46.8 & 54.9 & 49.6 & 57.0 & 81.2 & 56.6 & 59.4 & 77.4 \\
    CLIPort~\citep{shridhar2022cliport} & 63.2 & 78.3 & 88.2 & 28.8 & 64.2 & 71.4 & 37.9 & 52.6 & 80.1 & 45.9 & 41.7 & 49.6 & 62.0 & 62.1 & 77.1 & 49.5 & 50.3 & 60.0 \\

    CLIPort\textbf{-multi}~\citep{shridhar2022cliport} & 60.3 & 82.9 & 81.4 & 42.4 & 53.7 & 54.3 & 76.6 & 84.3 & 77.0 & 50.4 & 58.7 & 47.6 & 79.0 & 88.0 & 88.6 & 79.9 & 85.6 & 73.8\\
    MOKA~\citep{liu2024moka} & \multicolumn{3}{c}{-----------16.3-----------} & \multicolumn{3}{c}{-----------20.6-----------} & \multicolumn{3}{c}{-----------32.7-----------} & \multicolumn{3}{c}{-----------41.2-----------} & \multicolumn{3}{c}{-----------36.1-----------} & \multicolumn{3}{c}{-----------40.7-----------} \\
    \midrule
    GEM (ours) &79.6 & 86.7 & 91.8 & 67.3 & 71.4 & \textbf{78.2} & 76.2 & 85.8 & 89.7 & 69.2 & \textbf{79.8} & \textbf{86.0} & 86.6 & 85.1 & \textbf{94.2} & 78.1 & 71.9 & 82.3 \\
    GEM\textbf{-multi} (ours) & \textbf{90.6} & \textbf{90.7} & \textbf{93.8} & \textbf{73.8} & \textbf{78.2} & \textbf{78.2} & \textbf{93.7} & \textbf{91.0} & \textbf{90.3} & \textbf{86.3} & 79.7 & 75.7 & \textbf{94.5} & \textbf{93.1} & \textbf{94.2} & \textbf{89.7} & \textbf{90.9} & \textbf{88.5}\\
    \midrule
    
    & \multicolumn{3}{c}{\begin{tabular}[c]{@{}c@{}}stack-block-pyramid\\seq-seen-colors\end{tabular}}    & \multicolumn{3}{c}{\begin{tabular}[c]{@{}c@{}}stack-block-pyramid\\seq-unseen-colors\end{tabular}}    & \multicolumn{3}{c}{\begin{tabular}[c]{@{}c@{}}separating-piles\\seen-colors\end{tabular}} & 
    \multicolumn{3}{c}{\begin{tabular}[c]{@{}c@{}}separating-piles\\unseen-colors\end{tabular}} & 
    \multicolumn{3}{c}{\begin{tabular}[c]{@{}c@{}}towers-of-hanoi\\seq-seen-colors\end{tabular}} & 
    \multicolumn{3}{c}{\begin{tabular}[c]{@{}c@{}}towers-of-hanoi\\seq-unseen-colors\end{tabular}}  \\ [-2pt]
    
    \cmidrule(lr){2-4} \cmidrule(lr){5-7} \cmidrule(lr){8-10} \cmidrule(lr){11-13} \cmidrule(lr){14-16} \cmidrule(lr){17-19} \cmidrule(lr){20-22}
        Model &  \multicolumn{1}{c}{10}  & 20   & 100 & \multicolumn{1}{c}{10}  & 20   & 100 & 10 & 20 & 100 &  10 & 20 & 100 & 10  &  20 & 100 & 10 & 20 & 100 & \\
 \midrule

    Transporter-Lan~\citep{zeng2021transporter} & 52.0 & 72.7 & 94.3 & 18.0 & 26.0 &  17.0 & 40.0 & 60.0 & 92.0 & 56.0 & 73.8 & 52.3 & 81.1 & 88.6 & 95.7 & 43.4 & 48.3 & 60.0 
    \\
    CLIPort~\citep{shridhar2022cliport} & 22.8 & 39.5 & 50.5 & 21.8 & 19.2 & 27.7 & 53.1 & 56.0 & 74.8 & 56.4 & 66.0 & 72.5 & 75.1 & 75.0 & 91.1 & 57.6 & 47.3 & \textbf{99.4} \\
    CLIPort\textbf{-multi}~\citep{shridhar2022cliport} & 74.7 & 87.7 & 93.3 & 45.7 & 28.3 & 33.0 & 59.7 & 72.2 & 75.0 & 67.8 & 65.2 & 58.8 & 78.3 & 95.4 & 97.4 & 60.3 & 69.4 & 69.7\\
    MOKA~\citep{liu2024moka} & \multicolumn{3}{c}{-----------0-----------} & \multicolumn{3}{c}{-----------0-----------} & \multicolumn{3}{c}{-----------11.4-----------} & \multicolumn{3}{c}{-----------14.8-----------} & \multicolumn{3}{c}{-----------0-----------} & \multicolumn{3}{c}{-----------0-----------} \\
    \midrule
     GEM (ours) &70.7 & 82.7 & \textbf{96.3} & 59.3 & 73.7 & \textbf{84.3} & 82.3 & 75.4 & 78.8 & 60.0 & 91.8 & 96.6 & 88.3 & 93.4 & \textbf{100} & 83.1 & 87.7 & 98.0 \\
    GEM\textbf{-multi} (ours) & \textbf{94.3} & \textbf{95.3} & 95.0 & \textbf{76.0} & \textbf{89.3} & 78.7 & \textbf{94.2} & \textbf{96.2} & \textbf{92.0} & \textbf{89.0} & \textbf{97.6} & \textbf{96.6} & \textbf{96.3} & \textbf{99.4} & 98.9 & \textbf{93.4} & \textbf{98.0} & 97.1\\
    \midrule
    & \multicolumn{3}{c}{align-rope}    & \multicolumn{3}{c}{packing-unseen-shapes}    & \multicolumn{3}{c}{\begin{tabular}[c]{@{}c@{}}assembling-kits-seq\\seen-colors\end{tabular}} & 
    \multicolumn{3}{c}{\begin{tabular}[c]{@{}c@{}}assembling-kits-seq\\unseen-colors\end{tabular}} & 
    \multicolumn{3}{c}{\begin{tabular}[c]{@{}c@{}}put-blocks-in-bowls\\seen-colors\end{tabular}} & 
    \multicolumn{3}{c}{\begin{tabular}[c]{@{}c@{}}put-blocks-in-bowls\\unseen-colors\end{tabular}}  \\ [-2pt]
    
    \cmidrule(lr){2-4} \cmidrule(lr){5-7} \cmidrule(lr){8-10} \cmidrule(lr){11-13} \cmidrule(lr){14-16} \cmidrule(lr){17-19} \cmidrule(lr){20-22}
        Model &  10  & 20   & 100 & 10  & 20   & 100 & 10 & 20 & 100 &  10 & 20 & 100 & 10  &  20 & 100 & 10 & 20 & 100 & \\
 \midrule

    Transporter-Lan~\citep{zeng2021transporter} & 11.5 & 33.7 & 72.4 & 24.0 & 26.0 & 30.0 & 26.4 & 39.2 & 58.4 & 20.0 & 24.8 & 23.6 & 42.7 & 68.7 & 86.3 & 12.0 & 17.0 & 36.0 \\
    CLIPort~\citep{shridhar2022cliport} & 30.0 & 16.9 & 51.5 & 29.0 & 24.0 & 34.0 & 17.8 & 24.8 & 39.4 & 16.6 & 20.6 & 36.6 & 37.2 & 55.6 & 92.7 & 50.8 & 41.7 & 51.8 \\
    
    CLIPort\textbf{-multi}~\citep{shridhar2022cliport} & 39.7 & 42.4 & 40.8 & 52.0 & 46.0 & 52.0 & 28.8 & 42.8 & 32.0 & 28.4 & 27.2 & 18.8 & 84.0 & 96.0 & 98.0 & 38.7 & 48.0 & 44.0\\
    MOKA~\citep{liu2024moka} & \multicolumn{3}{c}{-----------2.4-----------} & \multicolumn{3}{c}{-----------4.0-----------} & \multicolumn{3}{c}{-----------0-----------} & \multicolumn{3}{c}{-----------0-----------} & \multicolumn{3}{c}{-----------22.5-----------} & \multicolumn{3}{c}{-----------17.5-----------}\\
    \midrule
    GEM (ours) & 31.6 & 38.6 & \textbf{69.0} &  54.0 & 44.0 & \textbf{52.0} & 42.8 & 47.2 & \textbf{62.4} & 34.4 & 40.0 & \textbf{62.8}& 94.0 & 98.3& \textbf{100} & 87.7 & 92.0 & 94.3 \\
    GEM\textbf{-multi} (ours)& \textbf{62.6} & \textbf{59.6} & 58.6 & \textbf{60.0} & \textbf{50.0} & \textbf{52.0} & \textbf{55.6} & \textbf{62.0} & 56.8 & \textbf{53.2} & \textbf{58.0} & 46.4 & \textbf{100} & \textbf{100} & \textbf{100} & \textbf{95.3} & \textbf{97.0} & \textbf{97.0}\\
    
    \bottomrule
    \end{tabular}
    \caption{
    \small
    \textbf{Performance Comparisons on CLIPort Benchmark Tasks (\%)} on 50 testing episodes. \{10, 20, 100\} denotes the number of demonstrations used in training. ``\textbf{-multi}'' denotes multi-task models where they are trained on all tasks and evaluated separately. }
    \label{tab:sim_result}
    \vspace{-18pt}
\end{table*}

\section{Experiments}
We design the following experiments to answer four questions:
\begin{itemize}
    \item How is our data efficiency and performance on seen and unseen scenarios compared with few-shot learning methods~\cite{shridhar2022cliport, zeng2021transporter}, large transformer-based models~\cite{jiang2022vima}, and zero-shot methods~\cite{liu2024moka}? --- Section~\ref{sec:simulation}
    \item How is the real-world performance of our method on different physical robot platforms? --- Section~\ref{sec:real-world}
    \item How is our robustness against unseen poses and language \& objects versus large VLA models~\cite{kim2024openvla} with few-shot fine-tuning? --- Section~\ref{sec:robustness}
    \item How does each component in our method contribute to the final performance? --- Section~\ref{sec:ablation}
\end{itemize}

\subsection{Simulation Experiments}
\label{sec:simulation}
\textbf{Tasks \& Baselines}
For simulation tasks, we use 18 tasks provided by CLIPort Benchmark~\citep{shridhar2022cliport}, including multi-step tasks (\textit{block-in-bowl}, \textit{packing-box/google-pairs/objects}), long-horizon (\textit{stack-block-pyramid-seq}, \textit{towers-of-hanoi}), deformable object (\textit{align-rope}, \textit{separating-piles}), novel color/object tasks (\textit{unseen} tasks). The metric is in the range of 0 (failure) to 100 (success). Partial rewards are calculated in multi-step tasks. For instance, in \textit{pushing-colored-piles}, pushing 10 piles out of 50 into the correct zone will be credited $\frac{10}{50}\times 100\%$ rewards. For baselines, we compare our method with four strong baselines: 
Transporter~\citep{zeng2021transporter}, CLIPort~\citep{shridhar2022cliport}, VIMA~\citep{jiang2022vima}, MOKA~\cite{liu2024moka}.

\begin{figure}[t]
\vspace{5pt}
    \centering
    \includegraphics[width=\linewidth]{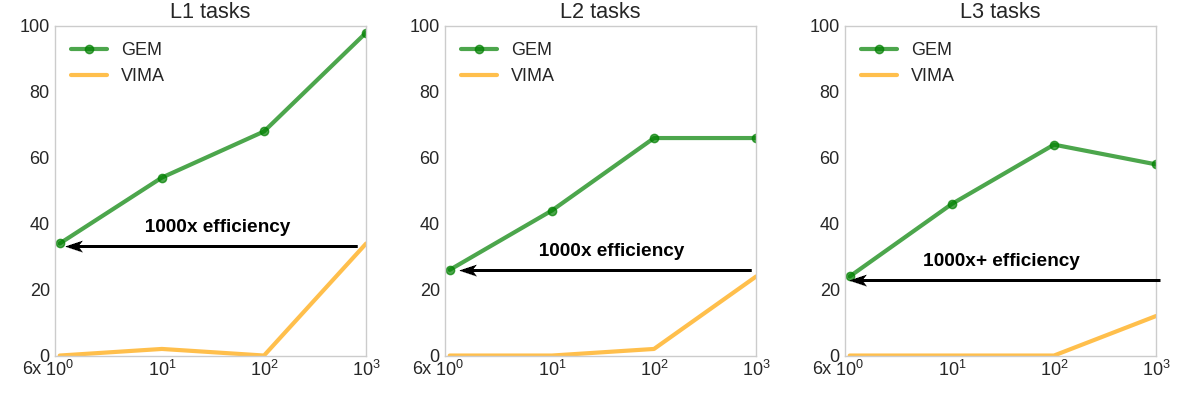}
    \caption{
    \small
    \textbf{Performance Comparisons on VIMABench~\citep{jiang2022vima}.} X-axis and y-axis represent the number of demonstrations for training and task success rate during evaluation in the \textit{visual\_manipulation} task. 
    }
    \label{fig:vima-gem-sim}
    \vspace{-18pt}
\end{figure}

\textbf{Comparing with Data-efficient CNN-based Methods:}
Transporter~\cite{zeng2021transporter} and CLIPort~\cite{shridhar2022cliport} are data-efficient methods based on convolutional neural networks for pick-and-place tasks. Transporter-Lan is a modified Transporter~\cite{zeng2021transporter} to take language instructions, where we concat the text embedding onto the bottleneck of UNet. In Table~\ref{tab:sim_result}, we report the performance on CLIPort Benchmark~\citep{shridhar2022cliport}. The best performance is highlighted in bold in each column. Several conclusions can be drawn from Table~\ref{tab:sim_result}.
\textbf{(1)} GEM outperforms the baselines in all the tasks by a significantly large margin.
For example, in task \textit{separating-piles-unseen-colors}, our method gets $97.6\%$ success rate with 20 demos while the best baseline only achieves $66.0\%$.
\textbf{(2)} GEM is more sample efficient compared with the baseline. Trained with 10 demos, it can outperform the baselines with 20 and 100 demos on 10 out of 18 tasks. In \textit{stack-block-pyramid-seq-seen-colors}, our method trained with 10 demos gets $70.7\%$ while CLIPort only gets $50.5\%$ trained with 100 demos.
\textbf{(3)} GEM demonstrates strong zero-shot learning ability. The performance gap between GEM and the baselines becomes larger with unseen objects. In \textit{put-blocks-in-bowls} with 100 demos, the difference between GEM and CLIPort increases from $\Delta 7.3\%$ to $\Delta 42.5\%$ on unseen colors. \textbf{(4)} GEM is capable of learning a multi-task policy from diverse datasets. As shown in multi-task results, GEM-multi performs best on 41 out of 54 evaluation cases.

\textbf{Comparing with Zero-shot VLM Methods:} In Table~\ref{tab:sim_result}, we also compare with MOKA~\cite{liu2024moka}, a state-of-the-art zero-shot method that is based on GPT4~\cite{openai2023gpt4} and SAM~\cite{kirillov2023segment} without training on any robot data. Table~\ref{tab:sim_result} shows that our method trained with 10 demonstrations achieves, on average, 43.67 \% better than the baseline across all tasks. The large performance gap indicates that (1) there is a large gap between LLM/VLM data and task-specific robot data, and (2) it is vital to do few-shot learning to align pre-trained relevancies and post-trained robotic features.

\textbf{Comparing with Transformer-based Methods:} VIMA~\citep{jiang2022vima} is a learning-based approach that is based on pretrained transformer architecture~\cite{vaswani17attention}. In Figure~\ref{fig:vima-gem-sim}, we compare GEM with VIMA~\citep{jiang2022vima} on VIMABench~\citep{jiang2022vima}, where ours achieves the same performance with 6 demos comparing VIMA with 6k demos.

\subsection{Real-world Experiments}
\label{sec:real-world}
For real-world experiments, we evaluate the few-shot and zero-shot learning ability of our model on a series of challenging open-vocabulary manipulation tasks. 

\begin{figure}[t]
    \centering
     \vspace{8pt}
    \begin{subfigure}[b]{0.45\linewidth}
         \centering
        \includegraphics[width=\textwidth]{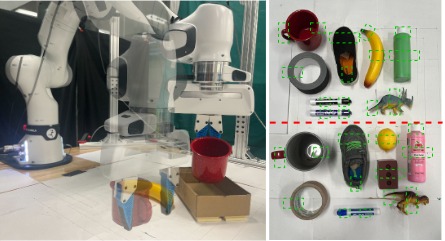}
         \caption{pick-object-part-in-box}
     \end{subfigure}
     \begin{subfigure}[b]{0.45\linewidth}
         \centering
        \includegraphics[width=\textwidth]{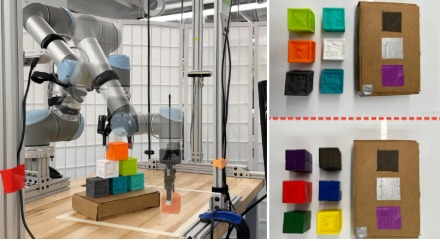}
          \caption{stack-block-pyramid}
     \end{subfigure}
     
     \begin{subfigure}[b]{0.45\linewidth}
        \includegraphics[width=\textwidth]{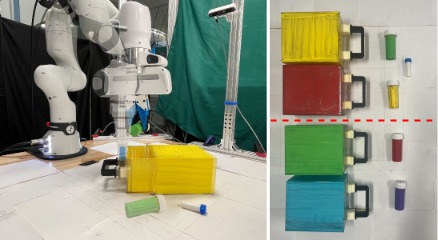}
          \caption{pill-organizing}
          \vspace{2mm}
     \end{subfigure}
     \begin{subfigure}[b]{0.45\linewidth}
         \centering
        \includegraphics[width=\textwidth]{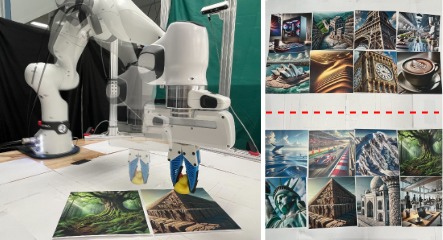}
          \caption{common-knowledge}
          \vspace{2mm}
     \end{subfigure}
     
    \vspace{-10pt}
    \caption{
    \small
    \textbf{Real-world Tabletop Tasks.} The transparent and solid arms show the intermediate actions. On each task's right side, it shows the seen/unseen objects on the upper/bottom sides.}

    \label{fig:tabletop-task}
    \vspace{-20pt}
\end{figure}

\textbf{Tabletop Tasks:} As shown in Figure~\ref{fig:tabletop-task}, we design four distinguishable tasks to demonstrate the effectiveness of our method on different levels. \textit{Pick-object-pat-in-box} demonstrates a key necessity of the few-shot learning ability of our method that we can learn to take fine-grained task-specific instructions like ``pick red mug \textbf{by its handle}''. \textit{Stack-block-pyramid} shows the method's capability of doing long-horizon sequential tasks with an $\SE(2)$ model in 3D space with z-axis heuristics. \textit{Pill-storaging} shows our method can do multi-task behaviors within one model beyond pick-and-place with simple primitive actions. \textit{Common-knowledge} demonstrates zero-shot recognition ability on common knowledge objects.

\begin{figure}[t]
    \vspace{7pt}
    \centering
    \begin{subfigure}[b]{0.21\linewidth}
         \centering
        \includegraphics[width=\textwidth]{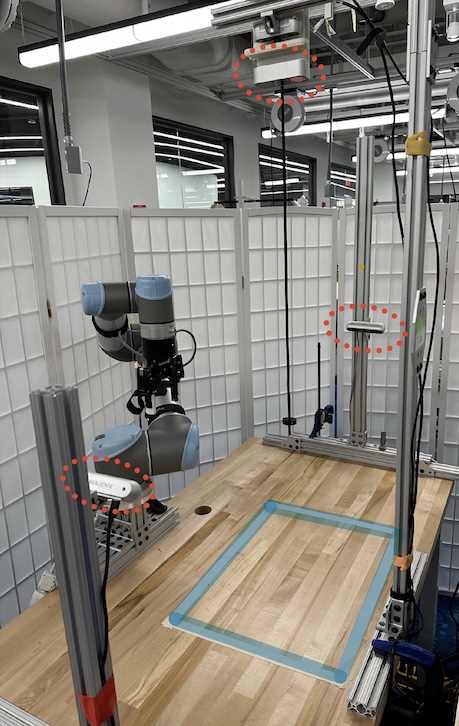}
         \caption{UR5}
     \end{subfigure}
     \begin{subfigure}[b]{0.23\linewidth}
         \centering
        \includegraphics[width=\textwidth]{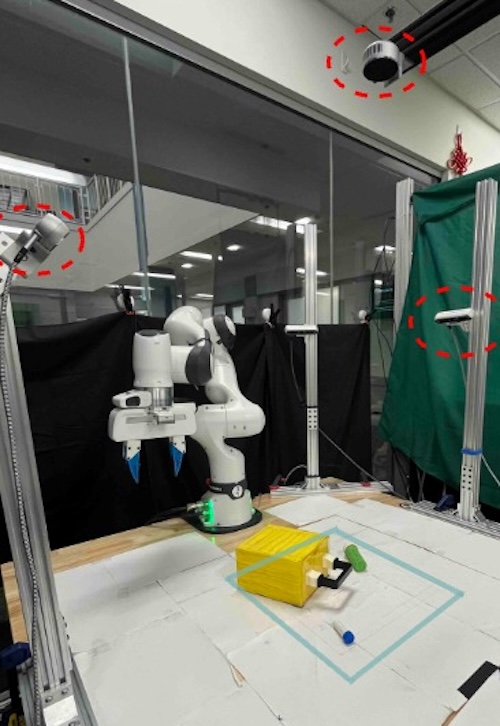}
         \caption{Franka}
     \end{subfigure}
     \begin{subfigure}[b]{0.35\linewidth}
         \centering
        \includegraphics[width=\textwidth]{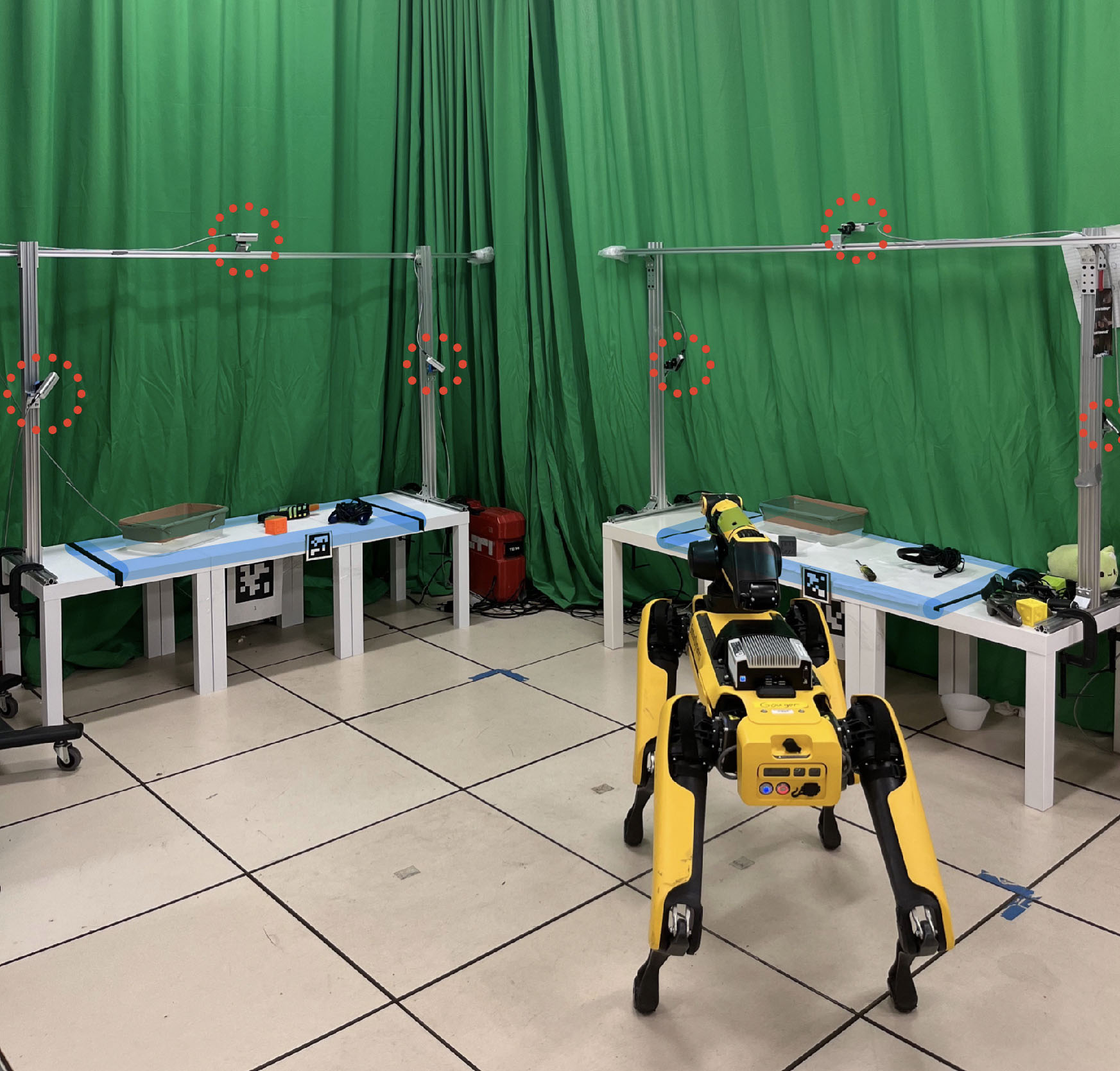}
         \caption{Spot}
     \end{subfigure}
     \vspace{-5pt}
    \caption{
    \small
    \textbf{Real-world Tabletop and Mobile Manipulation Setup.} Cameras are highlighted in red, and workspaces are labeled in blue.}
    \label{fig:settings}
    \vspace{-20pt}
\end{figure}

\textbf{Tabletop Results:} We evaluate our method in four tasks as shown in Figure~\ref{fig:tabletop-task} and report the results in Table~\ref{tab:tabletop_result}. Object sets can be found in Figure~\ref{fig:tabletop-task}. Our single-task method outperforms the baseline on all tasks up to a margin of $75.0\%$. On \textit{pick-object-part-in-box}, our method reaches $76.7\%$ success rate on seen objects while the baseline can only obtain $26.7\%$ success rate. Besides, it shows strong generalization ability on unseen colors and shapes, whereas the baseline fails to generalize well. The real-world experiments further prove the few-shot and zero-shot learning ability of GEM.

\begin{table}[h]
     \fontsize{9pt}{6pt}\selectfont
  \setlength\tabcolsep{4pt}
  \centering
  \scriptsize
    \begin{tabular}{@{}l*{5}{>{\centering\arraybackslash}p{10mm}@{}}}
    \toprule
    & \multicolumn{1}{c}{\begin{tabular}[c]{@{}c@{}}pick-object-\\part-in-box(5)\end{tabular}}     & \multicolumn{1}{c}{\begin{tabular}[c]{@{}c@{}}stack-\\block-pyramid(5)\end{tabular}}
    & \multicolumn{1}{c}{\begin{tabular}[c]{@{}c@{}}pill-\\storaging(5)\end{tabular}}
    & \multicolumn{1}{c}{\begin{tabular}[c]{@{}c@{}}common-\\knowledge(2)\\\end{tabular}}
      \\ [-2pt]
    
 \midrule

    CLIPort (seen) & 8/30 & 46/60 & 8/48 & 5/16  \\
    
    \textbf{GEM} (seen) & \textbf{23/30} & \textbf{56/60} & \textbf{35/48} & \textbf{16/16}  \\
    \midrule
    
    CLIPort (unseen) & 2/30 & 6/60 & 8/48 & 4/16 \\
    \textbf{GEM} (unseen) &\textbf{16/30} & \textbf{38/60} & \textbf{27/48} & \textbf{16/16}  \\

    \bottomrule
    \end{tabular}
    \caption{
    \small
    \textbf{Real-world Tabletop Performance}. The numbers next to task names indicate the n demonstrations we collect per training object per task. ``(seen)'' denotes that all objects are in-distribution, but their poses are randomly initialized during testing. ``(unseen)'' means all testing objects are novel. Denominators are the number of evaluation steps. Evaluation length differs due to different task lengths and numbers of objects.}
    \label{tab:tabletop_result}
    \vspace{-10pt}
\end{table}

\textbf{Mobile Manipulation Results:} 
To demonstrate the effectiveness of our method in an open-world setting with a large workspace, we evaluate GEM with Spot on a language-conditioned pick-place task. Pixel-based action space allows us to collect all the demos on one table, and the policy can generalize to a multi-table environment by using an action parameterization trick, where we concat observations in spatial dimension and unionize the output action map during inference time. Our model reaches $80\%$ success rate for seen objects and $50\%$ for unseen objects. 

\subsection{Generalization Robustness Analysis:}
\label{sec:robustness}
We evaluate the generalization robustness in terms of three aspects: unseen translations, unseen rotations, and novel objects with unseen language instructions. We compare with OpenVLA~\cite{kim2024openvla}, which is a large vision-language-action model pretrained on large-scale robotic datasets. We fine-tune OpenVLA on a few demonstrations and apply the same $\SE(2)$ data augmentation to guarantee fair comparison.

\begin{figure}[t]
\vspace{4pt}
     \begin{subfigure}[b]{\linewidth}
         \centering
        \includegraphics[width=\textwidth]{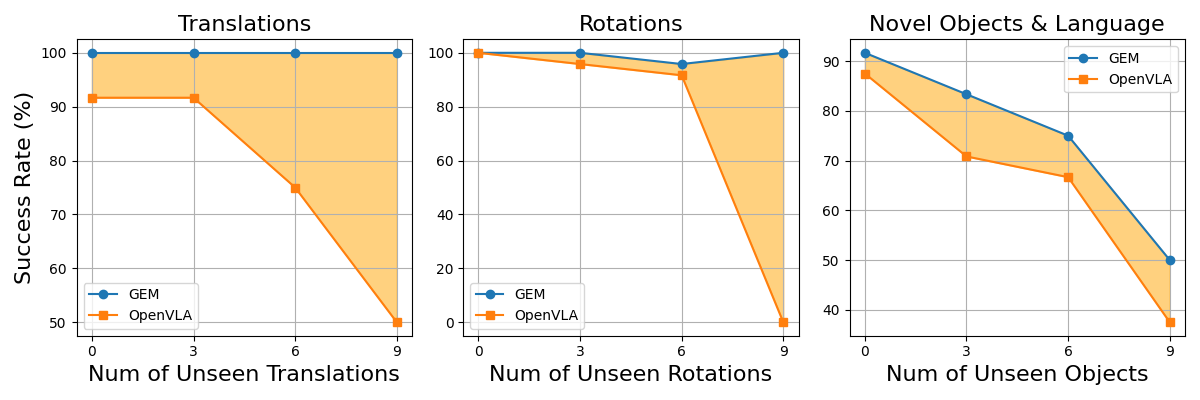}
     \end{subfigure}
    \caption{
    \small
    \textbf{Robustness Analysis} on Positional and Object Generalization. We highlight our robustness gain over the baseline in orange.
    }
    \vspace{-20pt}
    \label{fig:robustness}
\end{figure}

\textbf{Spatial Generalization Robustness:} We design a spatial generalization testbed to evaluate the translational and rotational generalization robustness. The task is to ``pick the red mug by handle and place into brown box'', where the robot needs to perform precise picking on the mug handle and put it into a box. We evenly collect demonstrations of 12 translations and rotations across the workspace separately. To test the robustness of translation generalization, we gradually decreased the amount of training data so that the number of unseen poses increases during evaluation. As shown in Figure~\ref{fig:robustness}, our method preserves robust performance when testing on more unseen poses. Meanwhile, the baseline suffers a severe performance drop when the data is limited.

\textbf{Novel Object Generalization Robustness:} We construct an object set that consists of 12 objects. The task is to ``pick \{OBJECT\} and place into brown box''. To test the robustness of novel objects and language instruction, we gradually decrease the number of training objects $n_{train}$ and test the models on the rest of the objects. During evaluation, the object orders are randomly shuffled so that all scenarios are unseen, even if they contain seen objects. OpenVLA~\cite{kim2024openvla} suffers more performance loss compared with our method when testing on more novel objects.

\begin{table}[h]
{
    \vspace{-10pt}
    \centering
    \scriptsize
    \setlength\tabcolsep{4pt}
    \begin{center}
    \begin{tabular}{c c c}
    \toprule
         &  pyramid-seen-100 & pyramid-unseen-100\\
         \midrule
    Ours     & \textbf{94.0} & \textbf{84.3}\\
    {w.o Relevancy Map} & 91.7 ($\downarrow$ 2.3) & 21.0 ($\downarrow$ 63.3) \\
    {w.o Multi-view Relevancy} & 63.0 ($\downarrow$ 31.0) & 34.3 ($\downarrow$ 50.0) \\
    {w.o Visual Relevancy Map} & 91.3 ($\downarrow$ 2.7) & 82.7 ($\downarrow$ 1.6)\\
    {w.o Steerable Kernel}& 7.7 ($\downarrow$ 84.0) & 1.7 ($\downarrow$ 82.6) \\
    {w.o Lan-cond Kernel} & 90.7 ($\downarrow$ 2.7) & 61.3 ($\downarrow$ 23.0) \\
    {CLIP $\rightarrow$ Ground DINO} & 83.7 ($\downarrow$ 10.3) & 21.7 ($\downarrow$ 41.6)\\
    \bottomrule
    \end{tabular}
    \end{center}
    \vspace{-10pt}
    \caption{
    \small
    \textbf{Ablation Study in Simulation.} Arrows indicate performance differences between ours and each variation.
    \label{tab:ablation}
    \vspace{-20pt}
    }
    }
\end{table}

\begin{table}[h]
{
\vspace{5pt}
\centering
\scriptsize
  \setlength\tabcolsep{2pt}
  \begin{center}
    \begin{tabular}{c cc}
    \toprule
         &  arrange-letter-to-word-seen & arrange-letter-to-word-unseen  \\
    \midrule
     Ours & \textbf{72.2} & \textbf{75.0}\\
     w.o Visual Relevancy Map & 42.2 ($\downarrow$ 30) & 68.8 ($\downarrow$ 6.2)\\
    \bottomrule
    \end{tabular}
    \end{center}
    \vspace{-10pt}
    \caption{
    \small
    \textbf{Ablation on Visual Relevancy Map in Real-world Tasks.} 5 demos are used for training. }
    \label{table:real-image-cond}
    }
    \vspace{-18pt}
\end{table}

\subsection{Ablation Study:}
\label{sec:ablation}
To investigate each component, we present a detailed ablation as shown in Table~\ref{tab:ablation} and Table~\ref{table:real-image-cond}. The main findings are: (1) Relevancy maps are crucial for novel object generalization, where the performance for pyramid-unseen drops by 63.3 without it. (2) Language steerable kernels are important for few-shot learning from limited demonstrations. (3) Visual relevancy maps are necessary for real-world tasks where the performance drops 6.2\%-30\% without it. Since real-world observations are often noisier, visual relevancy maps help suppress the noise. (4) Interestingly, multi-view observations can significantly affect performance. We hypothesize that side-view observations are closer to natural image views on which CLIP is trained. (5) The performance drops if we replace CLIP with an open-vocabulary object detector Grounding DINO~\cite{liu2023grounding}. It shows that relevancy maps are more scalable without the ``objectness'' inductive bias that object detectors often rely on.

\begin{table}[h]
  \setlength\tabcolsep{2pt}
  \centering
  \scriptsize
    \begin{tabular}{@{}l*{10}{>{\centering\arraybackslash}p{8mm}@{}}}
    \toprule
    & \multicolumn{3}{c}{block-in-bowl-seen}    & \multicolumn{3}{c}{block-in-bowl-unseen}    \\ 
    \midrule
        Number of Cameras &  1  & 2   & 3 & 1  & 2   & 3 \\
 \midrule
     
    Ours & \textbf{98.0} & \textbf{100.0} & \textbf{100.0} & \textbf{79.0} & \textbf{88.3} & \textbf{96.0}  \\
    CLIPort & 98.0 & 92.7 & 95.3 & 36.0 & 45.3 & 38.7 \\
    
    \bottomrule
    \end{tabular}
    \caption{
    \small
    \textbf{Ablation on Number of Camera Views.} 100 demos are used for training. Evaluated at 20,000 Steps.}
    \label{tab:num_cam}
    \vspace{-10pt}
\end{table}

The performance of our model gets affected, but still remains good performance given fewer camera views. As shown in Table~\ref{tab:num_cam}, reducing camera views from three to one, the performance drops by 2.0 and 17.0 in \textit{put-block-in-bowl-seen} and \textit{put-block-in-bowl-unseen}, respectively. Our hypothesis is that: (1) increased occlusion issues arise from fewer viewpoints, and (2) noisier saliency maps result from the absence of smoothing effects by multiple views. In contrast, CLIPort maintains stable performance given any number of cameras. However, our model still outperforms the baseline in all testing cases. In addition, the choice of viewpoint in the single-view setting is important. We find that the front-view camera that looks at the workspace from the opposite side of the robot works better than other views.

\section{Conclusion} 
\label{sec:conclusion}

In this work, we propose \textbf{G}rounded \textbf{E}quivariant \textbf{M}anipulation (\textbf{GEM}) that leverages pretrained textual-visual relevancy and the inherent symmetry in language-conditioned manipulation. Our method learns robust language-conditioned manipulation policies in a few-shot manner and shows a certain degree of novel object generalization ability. We validate the proposed method on simulated and real-world tasks against different baselines. One limitation is that our model needs predefined action primitives with heuristics like ``pick'', ``place'', ``open'', etc, limiting its ability to perform more complex tasks. Fortunately, LLMs~\cite{openai2023gpt4} and LLM-based methods~\cite{li2025hamster} provide convenient tools to parse and translate natural languages into high-level robot primitives. In the future, we will extend our method to (1) $\SE(3)$ action space without predefined primitives, and (2) improve local understanding~\cite{rawlekar2025disentangling, jing2024fineclip}. Worth mentioning, the language-conditioned symmetries we studied in Section~\ref{sec:method} are applicable for $\SE(3)$ space, which provides a solid foundation for extension using 3D convolution methods~\citep{james2022coarse, huang2024fourier}.

\footnotesize{
\bibliographystyle{IEEEtranN}
\bibliography{references}
}
\clearpage

\end{document}

%% file: math_commands.tex
\usepackage{amsmath,amsfonts,bm}

\def\eqref#1{equation~\ref{#1}}

\def\1{\bm{1}}

\DeclareMathAlphabet{\mathsfit}{\encodingdefault}{\sfdefault}{m}{sl}
\SetMathAlphabet{\mathsfit}{bold}{\encodingdefault}{\sfdefault}{bx}{n}

\DeclareMathOperator*{\argmax}{arg\,max}